\documentclass[10pt,twocolumn,letterpaper]{article}
\usepackage{multirow}
\usepackage{algorithm}
\usepackage{algorithmic}
\usepackage[algo2e, ruled]{algorithm2e}
\usepackage[pagenumbers]{iccv} 
\definecolor{iccvblue}{rgb}{0.21,0.49,0.74}
\usepackage[pagebackref,breaklinks,colorlinks,allcolors=iccvblue]{hyperref}
\newcommand{\methodshort}[1]{\textsc{Delta-DCT}}

\def\paperID{XXXX} 
\def\confName{ICCV}
\def\confYear{2025}

\title{
Seeing Delta Parameters as JPEG Images:\\Data-Free Delta Compression with Discrete Cosine Transform
}

\author{
Chenyu Huang$^{1}$\thanks{Equal Contribution} \quad Peng Ye$^{1,2,3 *}$ \quad Xiaohui Wang$^{1}$ \\ 
Shenghe Zheng$^{2,4}$
\quad Biqing Qi$^{2}$ \quad Lei Bai$^{2}$\quad Wanli Ouyang$^{2,3}$ \quad Tao Chen$^{1}$\thanks{Corresponding Author: \textit{eetchen@fudan.edu.cn}} \\ 
$^{1}$Fudan University \quad $^{2}$Shanghai Artificial Intelligence Laboratory \\ 
\quad $^{3}$The Chinese University of Hong Kong \quad
$^{4}$Harbin Institute of Technology \\
{\tt\small cyhuang24@m.fudan.edu.cn \quad eetchen@fudan.edu.cn}
}

\begin{document}
\maketitle

\begin{abstract}

With transformer-based models and the pretrain-finetune paradigm becoming mainstream, 
the high storage and deployment costs of individual finetuned models on multiple tasks pose critical challenges. 
Delta compression attempts to lower the costs by reducing the redundancy of delta parameters (i.e., the difference between the finetuned and pre-trained model weights).
However, existing methods usually face problems including data accessibility and training requirements.
To tackle this issue, we introduce Delta-DCT, the first data-free delta compression method inspired by classic JPEG image compression, leveraging the Discrete Cosine Transform (DCT).
We first 
(a) group delta parameters within a layer into patches. Then we 
(b) assess the importance of each patch and allocate them with different quantization bit-widths. Afterwards, we 
(c) convert these patches to the DCT domain and conduct quantization to each patch based on the allocated bit-width. The proposed Delta-DCT does not require any training or data calibration,
while achieving performance comparable to or even surpassing original finetuned models under 1-bit equivalent delta compression ratios on different kinds of models including: 
(1) recently-released LLMs of different sizes from 7B to 13B, 
(2) relatively smaller language models including RoBERTa and T5 models, 
(3) variants of vision transformer models, 
and (4) multi-modal BEiT-3 models.

\end{abstract}    

\section{Intrtoduction}

\begin{figure}[t]
    \centering
    \includegraphics[width=\linewidth]{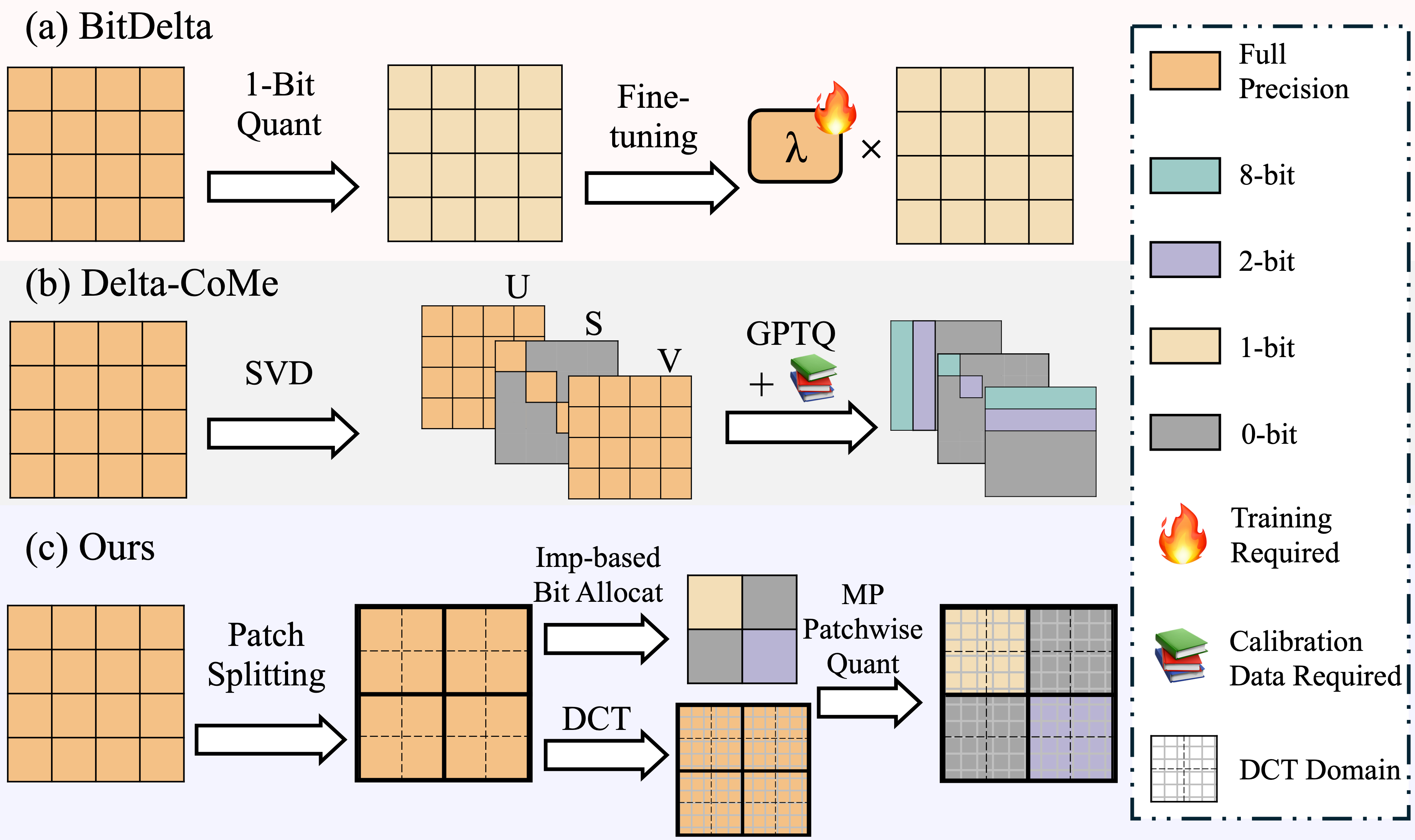}
    \caption{Compression methods comparison of (a) BitDelta~\cite{liu2024bitdelta}, which needs to finetune the scale factors; (b) Delta-CoMe~\cite{ping2024deltacome}, which employs calibration-data-requiring GPTQ~\cite{frantar2022optq}; and (c) Our \methodshort{}, which applies multi-precision patchwise quantization to the DCT converted patches via importance-based bit-width allocation, which requires no data or training.}
    \label{fig:motivation}
\end{figure}

With the proposal of transformer models~\cite{vaswani2017attention}, 
the pretrain-finetune paradigm has deeply affected the research fields including computer vision (CV)~\cite{dosovitskiy2020image, swin-transformer}, natural language processing (NLP)~\cite{devlin2018bert,radford2018improving}, and multi-modal learning~\cite{clip, liu2023llava}. Despite great success, there still exist some problems, including the models' performance being difficult to be optimal on different tasks simultaneously. Among the existing solutions, it is well known that using multiple task-specific models can theoretically achieve the best performance while resulting in massive storage and deployment costs.
Multi-task learning (MTL) partially solves this by reducing the number of task-specific models but inevitably introduces significant additional training cost~\cite{vandenhende2021multi, zhang2023uni3d, zhang2021survey}.
The recently-trending model merging method~\cite{wortsman2022model, TIESMerging, RegMean, tang2024merging, yang2024model} is a training-free alternative to MTL but its performance is unstable and usually deteriorates as the number of models to be merged increases.

Recently, with the discovery of the redundant properties of delta parameters~\cite{ding2023parameter, yu2023language} (i.e., the difference between the finetuned and pre-trained model weights),
delta compression~\cite{liu2024bitdelta, yao2023deltazip, ping2024deltacome} is proposed to resolve the above problem
by using a pre-trained model and several task-specific compressed delta parameters to replace multiple task-specific finetuned models.
By delta compression, the performance of individual finetuned models can be retained while the storage and deployment costs can be greatly reduced.
Specifically, BitDelta~\cite{liu2024bitdelta} quantizes the delta parameters to 1-bit and tunes the scaling factors of each layer through knowledge distillation.
However, tuning the scaling factors requires datasets and considerable computational resources, especially for Large Language Models (LLMs).
Delta-CoMe~\cite{ping2024deltacome} uses mixed precision quantization on the singular values of delta parameters to realize training-free delta compression of LLMs.
However, Delta-CoMe utilizes GPTQ~\cite{frantar2022optq}, a technique that demands data for calibration during inference, while such data may not be accessible due to privacy concerns.
Besides, existing methods mainly focus on compressing the delta parameters of LLMs while in actual scenarios, delta compression is required by end devices, on which smaller models are normally adopted including vision transformers~\cite{dosovitskiy2020image}, RoBERTa models~\cite{liu2019roberta}, and multimodal models. 

Similar to the high redundancy properties of the delta parameters, in the multimedia field, digital images contain considerable redundant information~\cite{dhawan2011review}, requiring huge amounts of storage space and large bandwidths for transmission~\cite{image_compression_tech}.
JPEG Compression~\cite{wallace1991jpeg} is a widely used digital image compression scheme, 
where its core is Discrete Cosine Transform (DCT)~\cite{1672377} and its procedures consist of two phases.
\textit{Encoding Phase}: 
i) Grouping the pixels within an image into $8\times8$ blocks;
ii) Block-wise DCT and quantization.
\textit{Decoding Phase}: 
Applying Inverse DCT (IDCT) to each block.
Generally, JPEG compression can significantly compress images while retaining most of the visual information. Moveover, it requires no additional data, and can be directly applied to most images.
Motivated by data-free and high-performance properties of JPEG compression, we explore compressing delta parameters in the DCT domain and propose \methodshort{}: (1) We first group delta parameters within a layer into patches; (2) Then we assess the importance of each patch and allocate them with different quantization bit-widths; (3) Afterwards, we convert these patches to the DCT domain and conduct quantization to each patch via the allocated bit-width. 
The schematic diagrams of different methods are shown in \cref{fig:motivation}. As comparisons, our method requires no dataset or training
during the whole process, realizing data-free delta compression.

We demonstrate the effectiveness of our \methodshort{} for Delta Compression by comprehensive experimental settings, including: 
(1) Recently-released LLMs of different sizes from 7B to 13B sizes, containing Llama-3.1~\cite{dubey2024llama3}, Deepseek-Math~\cite{deepseek-math}, finetuned on tasks covering Chat, Code, Math, and Safety. The proposed \methodshort{} can significantly lower the performance degradation caused by delta compression and even achieves better performance than the fine-tuned models, especially on code and safety tasks. (2) Relatively smaller language models including RoBERTa~\cite{liu2019roberta} and T5~\cite{raffel2020exploring} models finetuned on the GLUE~\cite{wang2018glue} benchmark. When applied to both these models, \methodshort{} outperforms the existing methods and may even exceed the performance of uncompressed finetuned models.
(3) Variants of vision transformer~\cite{dosovitskiy2020image} models finetuned on eight vision tasks. \methodshort{} shows the best average performance over existing methods, MTL, and uncompressed finetuned models.
(4) Multi-modal BEiT-3~\cite{beit3} models finetuned on three vision language tasks, which demonstrates the applicability of our \methodshort{} to multi-modal models.
(5) We also perform ablation studies to demonstrate the effectiveness of different procedures of the proposed method.

Our contributions can be summarized as:
\begin{itemize}
    \item Motivated by the classic JPEG compression, we explore the delta compression from the DCT domain for the first time. We first realize data-free delta compression and further reduce the performance degradation.

    \item We propose a framework based on compression in the DCT domain, named \methodshort{}, which consists of: i) we group delta parameters into patches and measure the importance degree of each patch for mixed bit-widths allocation; ii) we apply DCT to each patch and conduct mixed-precision patch-wise quantization, thus realizing data-free high-performance delta compression.

    \item The performance of the proposed \methodshort{} is evaluated across multiple classical and newly-established task settings on various model backbones, covering different language models (both LLM and relatively small LMs), vision models, and multi-modal models.
\end{itemize}

\begin{figure*}
    \centering
    \includegraphics[width=1\linewidth]{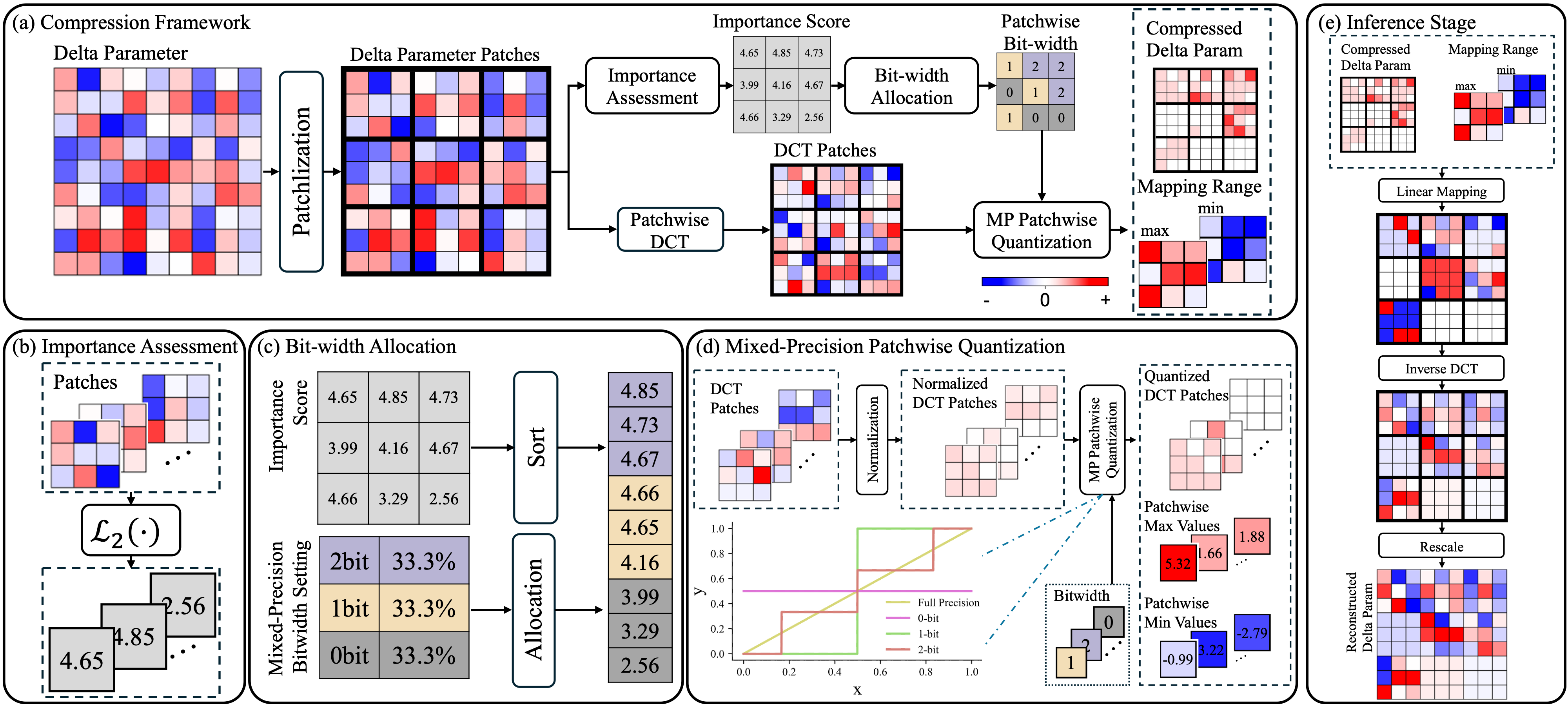}
    \caption{(a) The overview of the proposed data-free delta compression framework. We first divide the delta parameter into patches by patch size $p$. Then, we calculate the $\mathcal{L}_2$ Norm values of each patch as the importance assessment, as shown in (b). Based on the importance scores, we allocate different {bit-widths} to different patches via (c). Meanwhile, we apply Discrete Cosine Transform (DCT) to each patch, and further conduct mixed precision quantization to obtain the compressed data based on (d). During the inference stage, we first conduct linear mapping to the compressed delta parameters, and then inverse DCT (IDCT) and rescaling are conducted respectively for reconstruction, as shown in (e).
    }
    \label{fig:method-main}
    \vspace{-4pt}
\end{figure*}

\section{Related Work}

\textbf{Delta Compression} that compresses the delta parameters, i.e., the difference between the finetuned model parameters and the pre-trained model parameters, is becoming increasingly popular.
Compared to traditional model compression, delta parameters normally show extreme redundancy~\cite{yu2023language} and can be significantly compressed.
GPT-Zip~\cite{isik2023gpt} expands GPTQ~\cite{frantar2022optq} and successfully quantizes delta parameters to 2-bit.
DeltaZip~\cite{yao2023deltazip} combines quantization and sparsification to delta parameters to efficiently serve multiple LLMs.
BitDelta~\cite{liu2024bitdelta} quantizes the delta parameters to 1-bit and tunes the scaling factors of each layer through knowledge distillation.
Delta-CoMe~\cite{ping2024deltacome} utilizes mixed precision quantization based on the singular values of delta parameters to realize untrained delta compression of LLMs.
{However, all of these methods require either additional data or computational resources for calibration, whereas our \methodshort{} achieves both high performance and data-free compression. Besides, we first validate the performance of \methodshort{} on both LLMs and different kinds of other models, demonstrating its applicability.
}


\noindent 
\textbf{Model Merging} aims at combining the task-specific model weights and obtaining a multi-task model without joint training~\cite{yang2024model}.
Traditional static model merging methods usually generate a single model applicable to multiple tasks~\cite{wortsman2022model, RegMean, Task_Arithmetic, TIESMerging, ADAMerging}.
Recently, dynamic model merging, which means the merged model weights depend on the input samples, is getting increasingly popular due to their impressive performance~\cite{yang2024model, lu2024twinmerging, yang2024representation, zheng2024free}.
Compared to static methods, dynamic merging methods usually require additional modules to adapt the merged model to the target task~\cite{yang2024representation, huang2024emrmerging, ye2025mow}, thus improving performance. However, it still suffers from the sensitivity to the number of models to be merged and limited extensibility.
Compared to model merging, delta compression directly adapts the pre-trained base model to the target task using compressed delta parameters, thus showing strong extensibility and robustness.

\noindent 
\textbf{JPEG Compression} involves an encoding-decoding process~\cite{wallace1991jpeg} that utilizes DCT to convert image data into the frequency domain, followed by quantization and entropy coding. 
The compressed image is reconstructed using IDCT. 
JPEG compression is one of the most widely used techniques for image compression, particularly in large-scale datasets including ImageNet~\cite{deng2009imagenet} to store vast quantities of images efficiently.
Beyond its traditional use in image compression, JPEG compression has also been applied in image classification tasks to help models defend against adversarial attacks, improving their robustness~\cite{dziugaite2016study, guo2017countering}. 
However, to the best of our knowledge, little research has explored the use 
of JPEG compression in model weights to realize compression. 
Motivated by JPEG compression, in this paper, we propose \methodshort{}, which compresses delta parameters in the DCT domain and realizes data-free and high-performance delta compression.

\section{Method}
\label{sec:method}

\subsection{Preliminaries}
We follow the setting of existing delta compression methods~\cite{liu2024bitdelta, ping2024deltacome} and aim to compress the delta parameters on different tasks.
Specifically, when handling $N$ downstream tasks $\left[T_1, T_2, ..,T_N\right]$, instead of storing $N$ model weights $\left[\theta_1, \theta_2, ..,\theta_N\right]$ finetuned from the same pre-trained model $\theta_{pre}$, 
we store $\theta_{pre}$ and $\left[\delta_1, \delta_2,..,\delta_N\right]$, where $\delta_t$ is model $\theta_t$'s compressed delta parameter 
$\Delta_t = \theta_t-\theta_{pre}$,
$t\in N$. That is:

\begin{equation}
\label{eq:1}
\delta_t = \mathcal{C}\left(\Delta_t, \alpha\right),
\end{equation}
where $\mathcal{C}\left(\cdot\right)$ denotes the delta compression algorithm and $\alpha$ is the compression ratio.
Unlike traditional model compression, the redundancy within the delta parameters is more significant.
Existing works including BitDelta~\cite{liu2024bitdelta} and Delta-CoMe~\cite{ping2024deltacome} 
both quantize delta weights to 1-bit (or equivalent to 1-bit through mixed precision quantization),
which means $\alpha=\frac{1}{16}$ for FP16 or BF16 models and $\alpha=\frac{1}{32}$ for FP32 models in Eq.~\ref{eq:1}.

\begin{figure*}[t!]
    \vspace{-5pt}
    \centering
    \includegraphics[width=\linewidth]{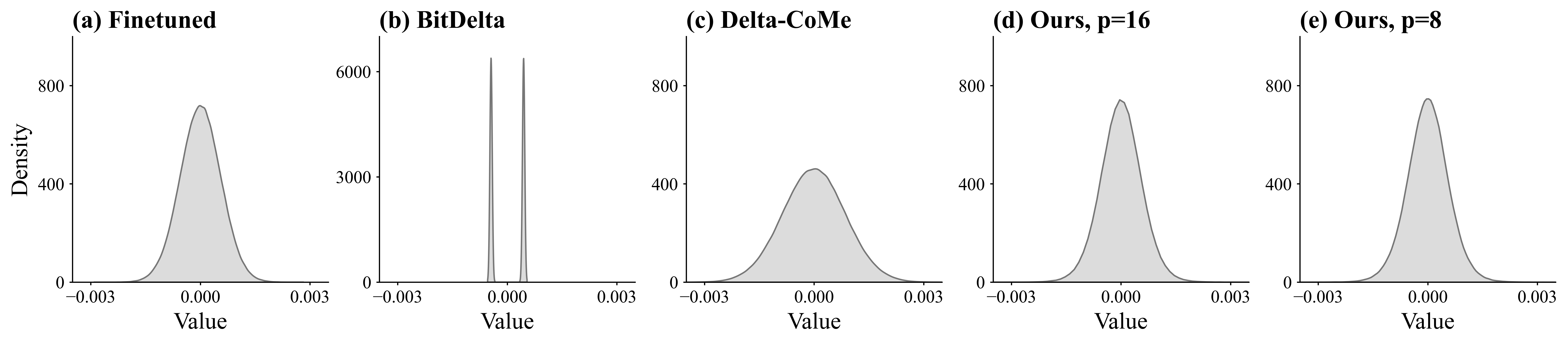}
    \caption{Visualization of delta parameter distribution of a layer in the (a) Finetuned model, (b) BitDelta~\cite{liu2024bitdelta} compressed model, (c) Delta-CoMe compressed model, and our \methodshort{} compressed model under the patch size setting of (d) $p=16$ and (e) $p=8$. 
    Existing methods usually cause obvious delta parameter distribution offsets while the delta parameter distribution of our \methodshort{} is almost the same as that of the finetuned model.}
    \label{fig:weigt_vis}
\end{figure*}

\subsection{\methodshort{}: Delta Compression in the DCT Domain}

In the digital image processing field, directly compressing the pixel values causes severe distortion of image quality.
In comparison, JPEG compression is conducted in the DCT domain, which can maintain most of the visible information of the image while requiring no additional data and can be directly applied to most images.
Therefore, we apply DCT to delta parameters before quantization.
Additionally, we design several modules, including patchwise importance assessment and bit-width allocation,
targeting high-performance data-free delta compression.
The detailed procedures of our \methodshort{} is shown in \cref{fig:method-main} and further illustrated as follows:

\noindent
\textbf{Compression stage:}

Overall, we first patchlize the delta parameters. Then we conduct an importance assessment of each patch. Based on the importance scores, we allocate the patchwise bit-widths to the patches.
Afterwards, we apply DCT to each patch and quantize them to the allocated bit-width.
Here we provide the detailed illustration of each procedure.

(1) \textit{Patchlization}. We dividing the delta parameters of layer $l$ of model $\theta_t$, $\Delta W^l$, into $p \times p$ patches:

\begin{equation}
\left[P_1^l, P_2^l, .., P_M^l\right]=\mathrm{Patch}\left(\Delta W^l,patchsize=p\right),
\end{equation}
where $\mathrm{Patch}\left(\cdot,patchsize=p\right)$ refers to patchlization with the patch size of $p$. $P^l_k$ is the $k$-th patch of layer $l$.

(2) \textit{Importance Assessment and Patchwise Bit-width Allocation}. We calculate $\mathcal{L}_2$ Norm of each patch as the importance score as follows:

\begin{equation}
\left[I_1^l,..I_M^l\right]=\left[\mathcal{L}_2\left( P_1^l \right),..,\mathcal{L}_2\left( P_M^l \right)\right], 
\end{equation}
where $I_k^l$ refers to the importance score of patch $P_k^l$.
Due to $\mathcal{L}_2$ Norm assesses magnitude changes,
the $\mathcal{L}_2$ Norm values of delta parameters measure the differences in parameter magnitudes, 
indicating the contribution of the delta parameters to the model's performance on the target task~\cite{tang2024merging}.
As a result, we consider patches with larger $\mathcal{L}_2$ Norm more important to the target task and apply higher quantization accuracy. 

Depending on the importance scores, we decide the quantization bit-width for each patch.
Specifically, we choose the pre-set quantization bit widths and corresponding ratios after sorting the importance scores to different patches $\left[B_1^l, .., B_M^l\right]$, where each bit width is within the mixed-precision quantization bit width setting $\left[b_1, ..,b_q\right]$ and ratios setting for each bit width $\left[r_1, ..,r_q\right]$.
The allocation principle follows that the patches with higher importance scores obtain the larger quantization bit-width.

(3) \textit{DCT and Compression}. We apply DCT and mixed-precision quantization to each patch, thus we obtain $\left[\hat{P_1^l}, \hat{P_2^l}, .., \hat{P_M^l}\right]$, where $\hat{P_k^l}=\mathrm{DCT}\left(P_k^l\right)$, $k\in M$.
Then we apply quantization to each patch:

\begin{equation}
\hat{p_k^l}=\mathrm{Quant}\left(\hat{P_k^l}, B_k^l\right),
\end{equation}
where $\mathrm{Quant}\left(\cdot,b\right)$ is the quantization function using bit-width of $b$. 
To realize data-free compression, we apply simple round quantization. 
Thus, we need to store the mapping range of each patch in addition. Specifically, the upper and lower bound, which are two float values.
Therefore, compared to existing methods, our \methodshort{} has storage overhead, which will be discussed in \cref{sec:overhead}.
It should also be noted that for 0-bit quantized patches, the compressed patch is filled with zeros, and the patch range boundaries are both set to the original patch's average value.





The algorithm flow of \methodshort{} compression can be found in \cref{sec:algo}.

\begin{table*}[ht]
\caption{The performance of different methods on large language models. The best results are bolded.
}
\label{llm_result}
\vspace{-6pt}
\begin{center}
\begin{small}
\resizebox{0.95\textwidth}{!}{
\begin{tabular}{l|cc|cc|cc|cc|c}
\toprule

\multirow{2}{*}{\bf Method} 
& \multicolumn{2}{c|}{\textsc{Tulu-3-8B-SFT}}
& \multicolumn{2}{c|}{\textsc{Llama-Guard-3-8B}}
& \multicolumn{2}{c|}{\textsc{OLMo-2-13B-SFT}}
& \multicolumn{2}{c|}{\textsc{NuminaMath-7B-TIR}}
& \multirow{2}{*}{\bf Avg}
\\

&  {\bf HumanEval+}
& {\bf MBPP+} 
& {\bf SafetyBench} 
& {\bf BBQ} 
& {\bf TruthfulQA} 
& {\bf Belebele} 
& {\bf GSM8K} 
& {\bf Math}
& \\

\midrule
{Pre-trained} 
&31.1&51.9&73.8& 40.9 &50.3 &37.0 &61.9 & 30.4 & 47.2
\\
{Finetuned}
&57.9 & 56.9&76.4&68.0  &54.3 & {40.0}&73.7 & 32.3 & 57.4

\\

\midrule

{BitDelta~\cite{liu2024bitdelta}}
&54.9 & 55.6
&\textbf{79.4}&62.6&53.6&39.8&74.2 &\textbf{36.6} &57.1

\\
{Delta-CoMe~\cite{ping2024deltacome}}
& 45.7 & 52.4
&76.1&66.4&\textbf{53.8} &38.4&73.7 & 33.0& 54.9

\\
\midrule
\textbf{\methodshort{} (Ours, p=64)}&
56.7 & \textbf{58.2}
&78.3&67.7&53.7 &39.9 & 73.2 & 33.7 &\textbf{57.7} 

\\
\textbf{\methodshort{} (Ours, p=32)}
&\textbf{58.5} & 57.7&78.6&{66.6}&53.2 &  \textbf{40.0}& 73.5 & 32.4 & 57.6
\\
\textbf{\methodshort{} (Ours, p=16)} 
&\textbf{58.5}
&55.9
&78.9&\textbf{69.4}&52.8 & 39.9 & \textbf{74.5} & 31.6 &\textbf{57.7}
\\

\bottomrule
\end{tabular}
}
\end{small}
\end{center}

\end{table*}

\noindent
\textbf{Inference stage:}

The inference stage of \methodshort{} mainly contains two procedures:

(1) \textit{IDCT}. Reconstructing the delta parameters of layer $l$ by applying IDCT to each patch.
\begin{equation}
\begin{split}
\Delta W^{l\prime}_{recon} &=\left[{P^{l\prime}_1}, .., {P_M^{l\prime}}\right] \\
&=\left[\mathrm{IDCT}\left(\hat{p^{l\prime}_1}\right), .., \mathrm{IDCT}\left(\hat{p^{l\prime}_M}\right)\right],
\end{split}
\end{equation}
where $\hat{P^{l\prime}_k}$ is the $k$-th reconstructed patch. 
$\Delta W^{l\prime}_{recon}$ is the initially reconstructed $l$-th delta parameter layer.

(2) \textit{Rescaling}. We follow BitDelta's~\cite{liu2024bitdelta} rescaling strategy and align the average absolute values of each delta parameter layer.
\begin{equation}
\Delta W^l_{recon}=\Delta W^{l\prime}_{recon} \cdot \frac{\sum_{i,j}{\left|\Delta W^l\left[ i,j\right]\right|}}{\sum_{i,j}{\left|\Delta W^{l\prime}_{recon}\left[ i,j\right]\right|}},
\end{equation}
where $\Delta W^l_{recon}$ is the final reconstructed delta parameter of layer $l$.

\begin{figure}[t]
    \centering
    \includegraphics[width=\linewidth]{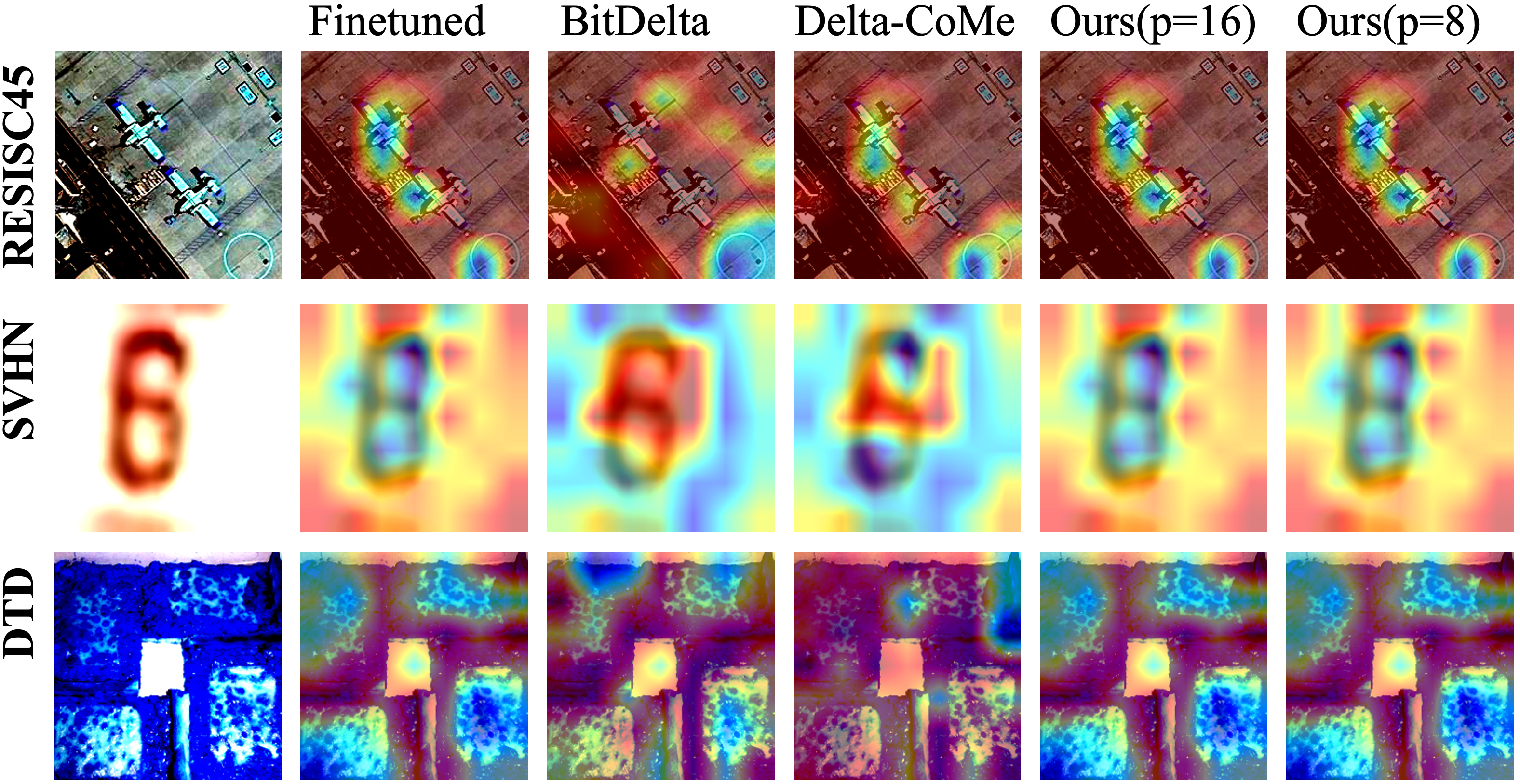}
    \caption{Grad-CAM visualization results of ViT-B/32 models compressed by different delta compression methods. Existing methods usually cause shifts in the focused areas while our \methodshort{} with different patch sizes consistently focuses on the area closest to the finetuned model.}
    \label{fig:cam}
\end{figure}

\begin{table*}[t]

\centering
\caption{The performance of different methods on T5-base models.}
\vspace{-4pt}
\label{tab:performance_t5_base} 
\resizebox{0.85\textwidth}{!}{
\begin{tabular}{l|cccccccc|c}
\toprule

{Methods}& CoLA&MNLI&MRPC&QNLI&QQP&RTE&SST2&STSB & \textbf{Avg} \\
\midrule
Pre-trained&69.13&56.45&76.23&88.45&82.12&80.14&91.17&62.19&75.74
\\
{Finetuned}  & 74.98&83.41&87.50&91.49&85.37&85.92&93.58&88.70&86.37  \\
\midrule
{BitDelta}~\cite{liu2024bitdelta} &  70.09&{83.64}&84.06&90.94&85.26&83.75&93.46&86.20&84.68\\

Delta-CoMe~\cite{ping2024deltacome}
& 74.88& 83.43 & 87.25 & 91.56 &{85.45} & {85.92} & {93.58} & 88.82 & 86.36
\\

\midrule

\textbf{\methodshort{} (Ours, p=16)} & 76.22 & 83.44 & {87.75} & {91.58} & 85.18 & {85.92} & {93.58} & 89.06& \textbf{86.59}

\\

\textbf{\methodshort{} (Ours, p=8)} &  {76.41} & 83.31 & 87.50 & {91.58} & 85.22 & 85.56 &{93.58} & {89.28} & 86.56

\\

\bottomrule
\end{tabular}
}

\end{table*}



\subsection{Reasons for effectiveness}

\noindent 
\textbf{Delta Parameter Distribution.}
In order to find out the reason for the effectiveness of the proposed \methodshort{},
we visualize the delta parameter distribution in \cref{fig:weigt_vis}.
We compare the reconstructed compressed delta parameters within a linear layer of the T5-Base~\cite{raffel2020exploring} model finetuned on CoLA~\cite{warstadt2019neural} from different compression methods.
Please check \cref{sec:exp_slm} for more experimental setting details.
The mixed precision quantization setting in our \methodshort{} is fully 1-bit quantization.
It can be seen that compared to other methods, the delta parameter distribution reconstructed by \methodshort{} is the closest to the finetuned model under the same bit width.
This may explain the improvement in the proposed method's performance.

\noindent 
\textbf{Grad-CAM Visualization.}
We provide visualization results of different delta compression methods on ViT-B/32 models using Grad-CAM~\cite{selvaraju2017grad} in \cref{fig:cam}.
The experimental settings can be found in \cref{sec:exp_vit}.
It can be seen that among all the delta compression methods, the visualization results of the proposed \methodshort{} are the closest to the uncompressed finetuned models, focusing on the most precise target.
The visualization results correspond to the quantized results, demonstrating that our \methodshort{} has the least affect on the focusing area of the model.

\section{Experiments}

\begin{table*}

\centering
\caption{The performance of different methods on ViT-B/32 models.}
\vspace{-6pt}
\label{tab:performance_vitbase32} 
\resizebox{0.85\textwidth}{!}{
\begin{tabular}{l|cccccccc|c}
\toprule

{Methods}& SUN397 & Cars &RESISC45 & EuroSAT & SVHN & GTSRB & MNIST & DTD  & \textbf{Avg Acc} \\
\midrule
Finetuned & 75.3  &  77.7  &  96.1  &  99.7  &  97.5  &  98.7  &  99.7  &  79.4 & 90.5   \\
{Traditional MTL} &    73.9  &  74.4  &  93.9  & 98.2    &  95.8  &  98.9   &  99.5   & 77.9 & 88.9  \\
\midrule




BitDelta~\cite{liu2024bitdelta}& 78.3 & 75.9 & 95.4& 99.4& 96.4 & 98.2 & 99.2 & 78.4 & 90.1\\
Delta-CoMe~\cite{ping2024deltacome}&78.6	&72.3	&{95.7}	&{99.8}&	{97.3}	&98.6&	{99.7}	&78.3&	90.0
\\

\midrule



\textbf{\methodshort{} (Ours, p=16)}& 
79.0&
78.4&
95.5 &
99.6 & 
97.1  &
98.8& 
99.6 &
79.1& 
90.9 \\


\textbf{\methodshort{} (Ours, p=8)}
&{79.3} &{78.8} & 95.6 & 99.7 & 97.2 & {98.9} & 99.6 &{79.3} & \textbf{91.1}
\\

\bottomrule
\end{tabular}
}

\end{table*}

\begin{table*}

\centering
\caption{The performance of different methods on ViT-L/14 models.}
\vspace{-6pt}
\label{tab:performance_vitlarge14} 
\resizebox{0.85\textwidth}{!}{
\begin{tabular}{l|cccccccc|c}
\toprule

{Methods}& SUN397 & Cars &RESISC45 & EuroSAT & SVHN & GTSRB & MNIST & DTD  & \textbf{Avg Acc} \\
\midrule
Finetuned  &  82.3  &  92.4  &  97.4  &  100  &  98.1  &  99.2  &  99.7  &  84.1  & 94.2   \\
{Traditional MTL}&80.8   &  90.6   &   96.3  & 96.3   & 97.6   & 99.1   &  99.6  &  84.4   & 93.5    \\
\midrule
BitDelta~\cite{liu2024bitdelta}&84.0&92.1 & 97.2 & 99.7 & 97.9 & 99.0 & 99.7 & 83.0 & 94.1 \\
Delta-CoMe~\cite{ping2024deltacome}&84.3&	92.1 & 	97.2	&99.6&	98.1	&99.1	&99.7&	83.6&	94.2\\
\midrule

\textbf{\methodshort{} (Ours, p=16)}&
84.6 & 92.3 & 97.3 & 99.7 & 98.0 & 99.3 & 99.7 & 84.0 & \textbf{94.4}\\
\textbf{\methodshort{} (Ours, p=8)}&84.7& 92.4 & 97.3 & 99.7 &98.0 & 99.3 & 99.7 & 84.0 & \textbf{94.4}\\

\bottomrule
\end{tabular}
}

\end{table*}

\begin{table*}

\centering
\caption{The performance of different methods on multi-modal BEiT-3 models.}
\label{tab:performance_beit} 
\resizebox{0.85\textwidth}{!}{
\begin{tabular}{lc|cccc|c|c}
\toprule

\multirow{2}{*}{Methods} &\multicolumn{1}{|c|}{Task}  & \multicolumn{4}{c|}{\textbf{COCO-Captioning}} & \textbf{NLVR2}& \textbf{ImageNet-1K}
\\
& \multicolumn{1}{|c|}{Metric}&  BLEU4($\uparrow$) &CIDEr($\uparrow$)&METEOR($\uparrow$) & ROUGE-L($\uparrow$)& Accuracy($\uparrow$) & Accuracy($\uparrow$)  
\\
\midrule

\multicolumn{2}{l|}{Finetuned} &  0.394 &1.337&0.311&0.601& 0.7765 &0.8537 
\\
\midrule
\multicolumn{2}{l|}{BitDelta~\cite{liu2024bitdelta}} & 
0.083&0.506&0.208&0.386&0.8232& 0.7987

\\



\midrule
\multicolumn{2}{l|}{\textbf{\methodshort{}(Ours,p=12)}} 
&0.382&1.306&0.307&0.595&0.8397&0.8235
\\
\multicolumn{2}{l|}{\textbf{\methodshort{}(Ours,p=8)}} 
&0.384 & 1.318 & 0.307 & 0.596&0.8381 &0.8252


\\

\bottomrule
\end{tabular}
}

\end{table*}

\label{sec:exp}
\textbf{Compared Methods.}
We compare the proposed \methodshort{} with:
(1)~Pretrained Models;
(2)~{Finetuned Models};
(3)~{Traditional MTL} (for ViTs), which uses datasets from all the tasks to train a single model jointly;
(4)~{BitDelta}~\cite{liu2024bitdelta}; and
(5)~{Delta-CoMe}~\cite{ping2024deltacome}. 
Please check \cref{sec:app_baseline} for a detailed description of BitDelta~\cite{liu2024bitdelta} and Delta-CoMe~\cite{ping2024deltacome}.
For a fair comparison, we validate the delta compression methods under no data settings.
Therefore, we use BitDelta-Initial (the initialized BitDelta without scale distillation) for BitDelta and apply round quantization instead of GPTQ for Delta-CoMe.
The bit-width for BitDelta is 1-bit and for Delta-CoMe, we follow the default setting, i.e., `8+3+2' triple-precision setting.
All the methods are evaluated on a single NVIDIA A100 GPU.

\noindent
\textbf{Hyper-parameter settings.}
We report the performance of our \methodshort{} under different settings of hyper-parameter $p$. 
Unless otherwise stated, in this section, the mixed-precision quantization is set to double-precision, that is, applying 2-bit quantization to 50\% patches and 0-bit (sparsification) to another 50\% patches, ensuring the compression ratio equivalent to 1-bit quantization, i.e., $\alpha=\frac{1}{16}$ for BF-16 LLMs and $\alpha=\frac{1}{32}$ for FP-32 models.
\cref{sec:ablation2} illustrates the reason for this mixed-precision quantization setting.

\subsection{Performance on LLMs}

\label{sec:exp_llm}

\noindent 
\textbf{Settings.}
LLMs with different architectures and sizes, including 7B, 8B, and 13B, are selected to evaluate the performance of the proposed method.
The checkpoints are all downloaded from huggingface~\cite{wolf2019huggingface} and their IDs are shown in \cref{app:llm_id}.
Eight tasks including HumanEval+~\cite{chen2021codex, evalplus}, MBPP+~\cite{austin2021program,evalperf}, TruthfulQA~\cite{lin-etal-2022-truthfulqa}, Belebele~\cite{bandarkar2023belebele},
GSM8K~\cite{cobbe2021trainingverifierssolvemath}, Minerva Math~\cite{2206.14858,hendrycksmath2021}, SafetyBench~\cite{zhang2023safetybench}, 
and BBQ~\cite{parrish2021bbq},
are applied to evaluate the LLMs, covering code, chat, math, and safety fields.

\noindent 
\textbf{Results.}
The results for LLMs are shown in Tab.~\ref{llm_result}. 
We observe that \methodshort{} performs nearly as well as the finetuned models on each task. 
On several tasks, particularly in code and safety fields, including HumanEval+, MBPP+, SafetyBench, BBQ, Math, and GSM8K, DELTA-JPEG even outperforms the finetuned models.
Compared to BitDelta~\cite{liu2024bitdelta} and Delta-CoMe~\cite{ping2024deltacome}, \methodshort{} demonstrates superior overall performance.
It especially excels in code generation tasks such as HumanEval+ and MBPP+, where it maintains near-perfect performance compared to fine-tuned models. In contrast, both BitDelta and Delta-CoMe have significant performance degradation in code generation tasks.
We also experiment with varying the patch size hyperparameter. \methodshort{} shows only minor fluctuations in performance, and its overall performance consistently outperforms other compression methods. 
These results highlight that \methodshort{} can achieve outstanding outcomes in LLMs while maintaining high efficiency and stability.

\subsection{Performance on other language models}
\label{sec:exp_slm}

\noindent 
\textbf{Settings.}
We follow the settings from FusionBench~\cite{tangFusionBenchComprehensiveBenchmark2024} and DARE~\cite{yu2023language}.
We validate the performance of T5-base~\cite{raffel2020exploring} models and RoBERTa-base~\cite{liu2019roberta} models on different tasks from GLUE~\cite{wang2018glue} benchmark, respectively CoLA~\cite{warstadt2019neural}, 
SST-2~\cite{socher2013recursive}, 
MRPC~\cite{dolan2005automatically},
STS-B~\cite{cer2017semeval},
QQP~\cite{iyer2017first},
MNLI~\cite{williams2017broad},
QNLI~\cite{rajpurkar2016squad}, and RTE~\cite{giampiccolo2007third}.
Following the existing settings, for T5-base models, STS-B is evaluated by Spearman’s $\rho$ and other tasks are evaluated by exact match accuracy.
The detailed settings and results for RoBERTa-base models are illustrated in \cref{sec:app-roberta}.

\noindent 
\textbf{Results.}
The results of T5-base models are shown in Tab.~\ref{tab:performance_t5_base}.
We observe that \methodshort{} achieves performance comparable to or even exceeding that of finetuned models.
When compared to BitDelta~\cite{liu2024bitdelta} and Delta-CoMe~\cite{ping2024deltacome}, \methodshort{} outperforms both methods on average scores, with particularly strong performance on specific tasks. For instance, \methodshort{} significantly outperforms other compression methods on CoLA task for the T5-base model. 
These results suggest that \methodshort{} can achieve promising results on not only LLMs, but also relatively small language models.

\subsection{Performance on vision models}
\label{sec:exp_vit}

\noindent 
\textbf{Settings.} 
We follow the settings from Task Arithmetic~\cite{Task_Arithmetic} and employ ViT-B/32 and ViT-L/14, two variants of CLIP~\cite{clip} models' visual encoders, as the pre-trained models.
The performance of each method is evaluated by eight image classification tasks, including SUN397~\cite{xiao2010sun},
Cars~\cite{krause20133d},
RESISC45~\cite{cheng2017remote},
EuroSAT~\cite{helber2019eurosat},
SVHN~\cite{yuval2011reading},
GTSRB~\cite{stallkamp2011german},
MNIST~\cite{lecun1998mnist}, and
DTD~\cite{cimpoi2014describing}.
All the datasets are evaluated by accuracy.
Additionally, we evaluate the performance of our method on ViT-B/16 models. Please check Appendix~\ref{sec:app-vit-b-16} for more details.

\noindent 
\textbf{Results.}
The results of ViT-B/32 models and ViT-L/14 models are respectively shown in Tab.~\ref{tab:performance_vitbase32} and \ref{tab:performance_vitlarge14}.
We observe that \methodshort{} performs well in vision models, consistently outperforming other compression methods and traditional multi-task learning (MTL) approaches. 
Furthermore, the average performance of \methodshort{} even surpasses individual finetuned models on both ViT-B/32 models and ViT-L/14 models, while other compression methods and MTL method all experience performance degradation.

\begin{table}

\centering
\caption{Compressing delta parameters of T5-Base models through \methodshort{} with and without DCT.}
\label{tab:ablation_dct} 
\resizebox{0.475\textwidth}{!}{
\begin{tabular}{l|cccccccc|c}
\toprule

{Methods}& CoLA&MNLI&MRPC&QNLI&QQP&RTE&SST2&STSB & \textbf{Avg} \\

\midrule

\multicolumn{7}{l}{\textbf{\methodshort{}(Ours, p=16)}}
\\
\midrule
w/ DCT & 76.22 & 83.44 & 87.75 & 91.58 & 85.18 & 85.92 & 93.58 & 89.06& 86.59

\\

w/o DCT& 69.32 & 82.72 & 80.15 & 90.39 & 84.67 & 83.03 & 92.66 & 78.84 & 82.72
\\
\midrule

\multicolumn{7}{l}{\textbf{\methodshort{}(Ours, p=8)}}
\\
\midrule
w/ DCT & 76.41 & 83.31 & 87.50 & 91.58 & 85.22 & 85.56 & 93.58 & 89.28 & 86.56 \\
w/o DCT&69.70 & 83.25 & 80.64 & 90.55 & 84.87 &83.39 & 92.89 & 83.84&83.64 \\
\bottomrule
\end{tabular}
}

\end{table}

\begin{figure*}
    \centering
    \includegraphics[width=0.98\linewidth]{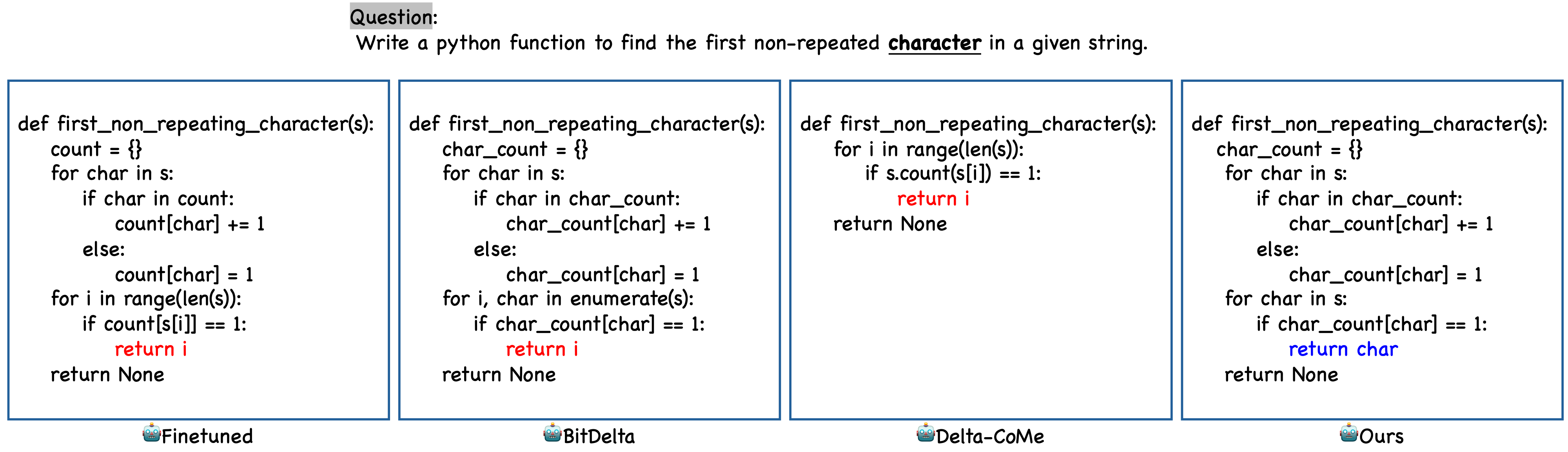}
    \caption{Case Study of a coding task for different delta compression methods. Only our \methodshort{} outputs the correct answer.
    }
    \label{fig:case}
\end{figure*}

\subsection{Performance on multi-modal models}

\noindent 
\textbf{Settings.}
We compress delta parameters on BEiT3-base~\cite{beit3} models finetuned on three datasets from different kinds of tasks, respectively ImageNet-1K~\cite{deng2009imagenet} (Image Classification), NLVR2~\cite{suhr2018corpus} (Visual Reasoning), and COCO Captioning~\cite{lin2014microsoft} (Image Captioning).
Among them, COCO Captioning is evaluated by BLEU4~\cite{papineni2002bleu}, CIDEr~\cite{vedantam2015cider},
METEOR~\cite{banerjee2005meteor},
and ROUGE-L~\cite{lin2004rouge}.
The other two tasks are evaluated by accuracy.

\noindent 
\textbf{Results.}
The results of BEiT3-base models are shown in Tab.~\ref{tab:performance_beit}.
We observe that \methodshort{} significantly outperforms BitDelta~\cite{liu2024bitdelta} in all the tasks and approaches the performance of individual models. In the COCO-Captioning task, \methodshort{} excels across all four metrics, closely matching the individual models, while BitDelta's performance is notably lower. In the NLVR2 task, \methodshort{} achieves an 8.14\% improvement in accuracy compared to the individual models. In the ImageNet-1K task, \methodshort{} outperforms BitDelta by 3.32\% in accuracy. Overall, \methodshort{} demonstrates strong performance across multi-modal tasks, highlighting its superiority and adaptability for various model types.

\begin{table}

\centering
\caption{Compressing delta parameters of T5-Base models using DCT delta parameter quantization and sparsification.}
\label{tab:ablation2} 
\resizebox{0.475\textwidth}{!}{
\begin{tabular}{l|cccccccc|c}
\toprule{Methods}& CoLA&MNLI&MRPC&QNLI&QQP&RTE&SST2&STSB & \textbf{Avg} \\

\midrule
\multicolumn{4}{l}{\textbf{Quantization}}
\\
\midrule

PTM&69.13&56.45&76.23&88.45&82.12&80.14&91.17&62.19&75.74
\\
1-bit & 74.88 & 79.28 & 86.76 & 92.26 & 83.42 & 83.75 & 92.78 & 87.67 & 85.10
\\
2-bit & 75.36 & 83.35 & 87.50 & 91.51 & 85.34 & 85.92 & 93.46 & 88.92 & 86.42
\\
4-bit& 74.98 & 83.41 & 87.50 & 91.51 & 85.37 & 85.92 & 93.58 & 88.70 & 86.37
\\
6-bit & 74.98& 83.41 & 87.50 & 91.49 & 85.36 & 85.92 & 93.58 & 88.70& 86.37
\\
8-bit & 74.98& 83.41 & 87.50 & 91.49 & 85.37 & 85.92 & 93.58 & 88.70& 86.37
\\
16-bit & 74.98& 83.41 & 87.50 & 91.49 & 85.37 & 85.92 & 93.58 & 88.70& 86.37
\\

32-bit & 74.98&83.41&87.50&91.49&85.37&85.92&93.58&88.70&86.37
\\
\midrule
\multicolumn{4}{l}{\textbf{Patch Sparsity}}
\\
\midrule

10\% & 75.35 & 83.27 & 87.99 & 91.51 & 85.33 & 85.92 & 93.46 & 89.05 & 86.49 
\\
20\% & 75.65 & 83.13 & 87.50 & 91.51 & 85.25 & 85.92 & 93.46 & 89.13 & 86.44 
\\
30\% & 76.70 & 83.12 & 88.24 & 91.54 & 85.13 & 85.92 & 93.58 & 89.22 & 86.68
\\
40\% & 77.56 & 82.98 & 87.99 & 91.62 & 85.02 & 85.92 & 93.69 & 89.27 & 86.76
\\
50\% & 77.76 & 82.74 & 88.48 & 91.63 & 84.86 & 85.92  & 93.58 & 88.90 & 86.74
\\
60\% & 77.09 & 82.27 & 87.99 & 91.71 & 84.58 & 85.92 & 93.35 & 88.48 & 86.42 
\\
70\% & 75.26 & 81.61 & 87.01 & 91.80 & 84.18 & 85.20 & 93.23 & 87.15 & 85.68
\\
80\% & 68.74 & 80.85 & 85.78 & 91.63 & 83.68 & 85.56 & 93.35 & 85.46 & 84.38
\\
90\% & 59.35 & 79.46& 78.68 & 91.34 & 82.88 & 87.00 & 93.69 & 82.53 & 81.87
\\
\midrule
\multicolumn{7}{l}{\textbf{\methodshort{} (Ours, 2-bit + 0-bit)}}

\\
\midrule

& 76.22 & 83.44 & 87.75 & 91.58 & 85.18 & 85.92 & 93.58 & 89.06& 86.59
\\
\bottomrule
\end{tabular}
}
\end{table}

\section{Discussion}
\subsection{Ablation Studies}

\textbf{Ablation on the DCT processing.}
To demonstrate the effectiveness of DCT, we compare the performance of our \methodshort{} with and without DCT in Tab.~\ref{tab:ablation_dct}, where \methodshort{} w/o DCT refers to dividing the delta parameters and applying multi-precision quantization without the DCT (and IDCT) process.
It can be seen that under both settings of hyper-parameter p, the DCT process can improve the performance significantly, demonstrating the effectiveness of DCT in our \methodshort{}.

\noindent
\textbf{Ablation on the fidelity of delta parameters.}
\label{sec:ablation2}
In order to thoroughly explore the redundancy of delta parameters in the DCT domain, we apply quantization with different {bit-widths} and sparsification with different patch sparse rates to the delta parameters in the DCT domain.
The results are shown in \cref{tab:ablation2} under the patch-size setting of $p=16$, demonstrating that delta parameters in the DCT domain can be significantly quantized or sparsified without obvious performance degradation.
Specifically, in the case of quantization settings of $\geq$ 2-bit or patch sparsification settings $\leq$ 60\%, the performance does not degrade and may even increase slightly.
This explains the mixed-precision setting of half 2-bit and half 0-bit in \cref{sec:exp}.

\subsection{Case Study}

We present a case study in \cref{fig:case}. 
The question is from MBPP~\cite{austin2021program}. We compare the answer from the finetuned model~\cite{lambert2024tulu3} and delta-compressed models by BitDelta~\cite{liu2024bitdelta}, Delta-CoMe~\cite{ping2024deltacome}, and our \methodshort{}. 
It can be seen that among all the methods, only our \methodshort{} outputs the correct answer, even outperforming the uncompressed finetuned model. 
Additional case studies could be found in \cref{sec:case_app}.

\begin{figure}[t]
    \centering
    \vspace{-4pt}
    \includegraphics[width=0.75\linewidth]{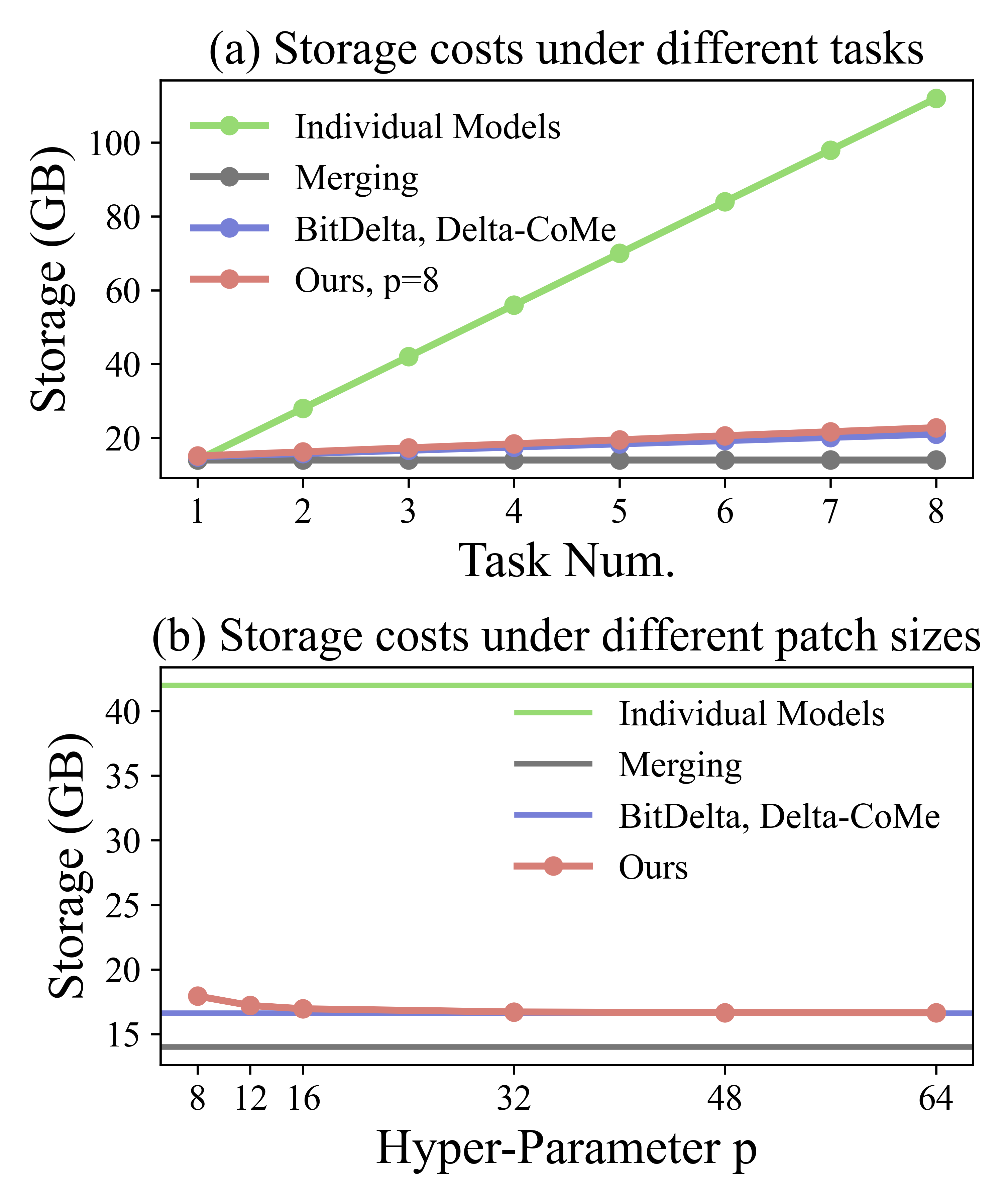}
    \caption{Storage costs of different methods (a) when applied to one to eight tasks, and (b) when applied to three tasks under different hyper-parameter settings of \methodshort{}'s patch size $p$. The storage costs of the proposed method are slightly greater than those of the existing methods but still significantly fewer than individual finetuned models.}
    \label{fig:overhead}
    \vspace{-5pt}
\end{figure}

\subsection{Overhead}
\label{sec:overhead}

Since we apply patch-wise mixed-precision quantization,
compared to layer-wise quantization, the proposed \methodshort{} has storage overhead. 
We present the overhead of our \methodshort{} by comparing the storage costs of \methodshort{}, individual finetuned models, a merged multi-task model, and existing delta compression methods, under the settings of different task numbers and different patch size $p$ in \cref{fig:overhead}(a) and \cref{fig:overhead}(b), respectively.
We assume that all the methods are applied to 7B LLMs.
It can be seen that the storage costs of the proposed method are slightly greater than those of the existing methods but still significantly fewer than individual finetuned models.

\section{Conclusion}

In this paper, we study data-free and high-performance delta compression.
Inspired by JPEG compression, we explore the effectiveness of compressing delta parameters in the DCT domain and propose \methodshort{}.
It reaches performance comparable to uncompressed models under the setting of 1-bit equivalent delta compression ratios.
The effectiveness of \methodshort{} is validated by comprehensive experiments on various benchmarks on both LLMs and other different kinds of models, covering vision, NLP, and multi-modal settings.

\section{Acknowledgement}

This work is supported by Shanghai Natural Science Foundation (No. 23ZR1402900), National Natural Science Foundation of China (No.62071127), National Key Research and Development Program of China (No.2022ZD0160101). The computations in this research were performed using the CFFF platform of Fudan University.

{
    \small
    \bibliographystyle{ieeenat_fullname}
    \bibliography{main}
}

\clearpage

\maketitlesupplementary
\appendix

\section{Existing Delta Compression Methods}
\label{sec:app_baseline}
\noindent
\textbf{BitDelta}~\cite{liu2024bitdelta} 
first obtains a binarized estimator by encoding the sign bits of the delta matrix $\Delta W$:
\begin{equation}
\hat{\Delta W} = \alpha \odot \mathrm{Sign}\left(\Delta W\right),
\end{equation}
where $\alpha$ is the scaling factor, the average of the absolute values of $\Delta W$. 
$\mathrm{Sign}\left(\cdot\right)$ obtains the sign matrix by $\mathrm{Sign}\left(\Delta W\right)=\mathrm{where}(\Delta W>0, +1, -1)$. 
Then the scaling factors are optimized for this objective:

\begin{equation}
    \arg\min_{{\boldsymbol{\alpha}}} \mathbb{E}_{x \in {X}}\left[ \left\| Z_{ft}\left(x\right)-Z_{bd}\left(x\right)\right\|^2 \right],
\end{equation}
where $X$ is the calibration dataset. $\boldsymbol{\alpha}$ denotes the collection of the scaling factors from all the layers.
$Z_{ft}$ and $Z_{bd}$ denote the finetuned model and the BitDelta reconstructed model, respectively.

\noindent
\textbf{Delta-CoMe}~\cite{ping2024deltacome} first use the SVD algorithm to decompose each delta matrix as follows:

\begin{equation}
\Delta W = USV^T,
\end{equation}
where $\Delta {W} \in \mathbb{R}^{h_{\mathrm{out}} \times h_{\mathrm{in}}}$, 
${U} \in \mathbb{R}^{h_{\mathrm{out}} \times h_{\mathrm{out}}}$, 
$S \in \mathbb{R}^{h_{\mathrm{out}} \times h_{\mathrm{in}}}$, 
${V} \in \mathbb{R}^{h_{\mathrm{in}} \times h_{\mathrm{in}}}$.
Delta-CoMe believes that the singular vectors associated with larger singular values have a greater impact on the approximation of the delta matrix $\Delta W$.
Based on the singular values Delta-CoMe applies varying quantization bits for different groups of singular vectors using GPTQ~\cite{frantar2022optq}.
\begin{equation}
\begin{split}
    \hat{V}[:, r_{\mathrm{begin}}:r_{\mathrm{end}}]^{\top} 
        &= \mathrm{Quant}_{k} \big( V[:, r_{\mathrm{begin}}:r_{\mathrm{end}}]^{\top},\ X \big), \\
    \hat{U}[:, r_{\mathrm{begin}}:r_{\mathrm{end}}] 
        &= \mathrm{Quant}_{k} \big( U[:, r_{\mathrm{begin}}:r_{\mathrm{end}}], \\
        S[r_{\mathrm{begin}}:&r_{\mathrm{end}},\ r_{\mathrm{begin}}:r_{\mathrm{end}}] \hat{V}[:, r_{\mathrm{begin}}:r_{\mathrm{end}}]^{\top} X \big),
\end{split}
\end{equation}
where $X$ is the input data and $\mathrm{Quant}_k$ denotes the $k$-bit GPTQ quantization.



\section{Algorithm flow of \methodshort{}}

\label{sec:algo}
We provide the algorithm flow of the compression procedure of \methodshort{} in \cref{algo:algo}.
The compression procedure does not require any data or training.
\begin{algorithm}[ht]
\caption{\methodshort{} Compression Procedure} 
\label{algo:algo} 
\KwIn{$L$-layer Finetuned model $\theta$, pre-trained model $\theta_{pre}$, where 
$\theta=[W^1,..,W^L]$. Patch-size $p$. Quantization bitwidth $\left[b_1,..b_q\right]$ and corresponding ratios $\left[r_1,..,r_q\right]$.
}
\KwOut{Compressed delta parameters $\hat{\delta}_{1..N}$.}

\For{$t ~\textbf{in} 1,...,N$}{

    $\triangleright$\ {\scriptsize Obtain delta parameters.} 
    
    $\Delta_t = \theta_t - \theta_{pre} = \left[\Delta W^1,..,\Delta W^L \right]$ 
    
    \For{$l ~\textbf{in} 1,...,L$}{

    $\left[P_1^l, P_2^l, .., P_M^l\right]=\mathrm{Patch}\left(W^l_t,patchsize=p\right)$

    $\triangleright$\ {\scriptsize Calculate the importance score for each patch.}
    
    $\left[I_1^l,..I_M^l\right]=\left[\mathcal{L}_2\left( P_1^l \right),..,\mathcal{L}_2\left( P_M^l \right)\right]$

    $\triangleright$\ {\scriptsize Allocate bitwidth for each patch based on importance scores.}
    
    $B_k^l = \mathcal{A}\left(I^l_{1..M},b_{1..q},r_{1..q}\right)$
    
    \For{$k \in \left[1,..,M\right]$}{

    $\triangleright$\ {\scriptsize Apply DCT and quantization.}
    
    $\hat{P_k^l}=\mathrm{DCT}\left(P_k^l\right)$

    $\hat{p_k^l}=\mathrm{Quant}\left(\hat{P_k^l}, B_k^l\right)$

   }

    $\hat{w^l_t}=\left[\hat{p_1^l}, \hat{p_2^l}, .., \hat{p_M^l}\right]$

    }
    $\triangleright$\ {\scriptsize Obtain the compressed delta parameters.}
    
    $\hat{\delta_t}=\left[\hat{w^1_t},..,\hat{w^L_t}\right]$
  
}



\end{algorithm}

\section{Model IDs for LLM checkpoints}
\label{app:llm_id}
Here in \cref{llm_settings}, we provide the model IDs for the LLM checkpoints utilized in \cref{sec:exp_llm}. 

\begin{table}[ht]
\caption{Selected LLM checkpoints for the examined tasks.}
\label{llm_settings}
\begin{center}
\begin{small}
\resizebox{0.475\textwidth}{!}{
\begin{tabular}{l|cc}
\toprule
\small{\textbf{Models}}
&Pre-trained&Finetuned
\\
\midrule
\small{Chat}
&
\textit{allenai/OLMo-2-1124-13B}~\cite{olmo20242}
&
\textit{allenai/OLMo-2-1124-13B-SFT}~\cite{olmo20242}
\\
\small{Code}
&
\textit{meta-llama/Llama-3.1-8B}~\cite{dubey2024llama3}
&
\textit{allenai/Llama-3.1-Tulu-3-8B-SFT}~\cite{lambert2024tulu3}
\\
\small{Math}
&
\textit{deepseek-ai/deepseek-math-7b-base}~\cite{deepseek-math}
&
\textit{AI-MO/NuminaMath-7B-TIR}~\cite{numina_math_7b}
\\
\small{Safety}
&
\textit{meta-llama/Llama-3.1-8B}~\cite{dubey2024llama3}
&
\textit{meta-llama/Llama-Guard-3-8B}~\cite{Inan2023LlamaGL}
\\













\bottomrule
\end{tabular}
}
\end{small}
\end{center}

\end{table}

\section{Additional Experiments}

\begin{figure*}[t]
    \centering
    \includegraphics[width=\linewidth]{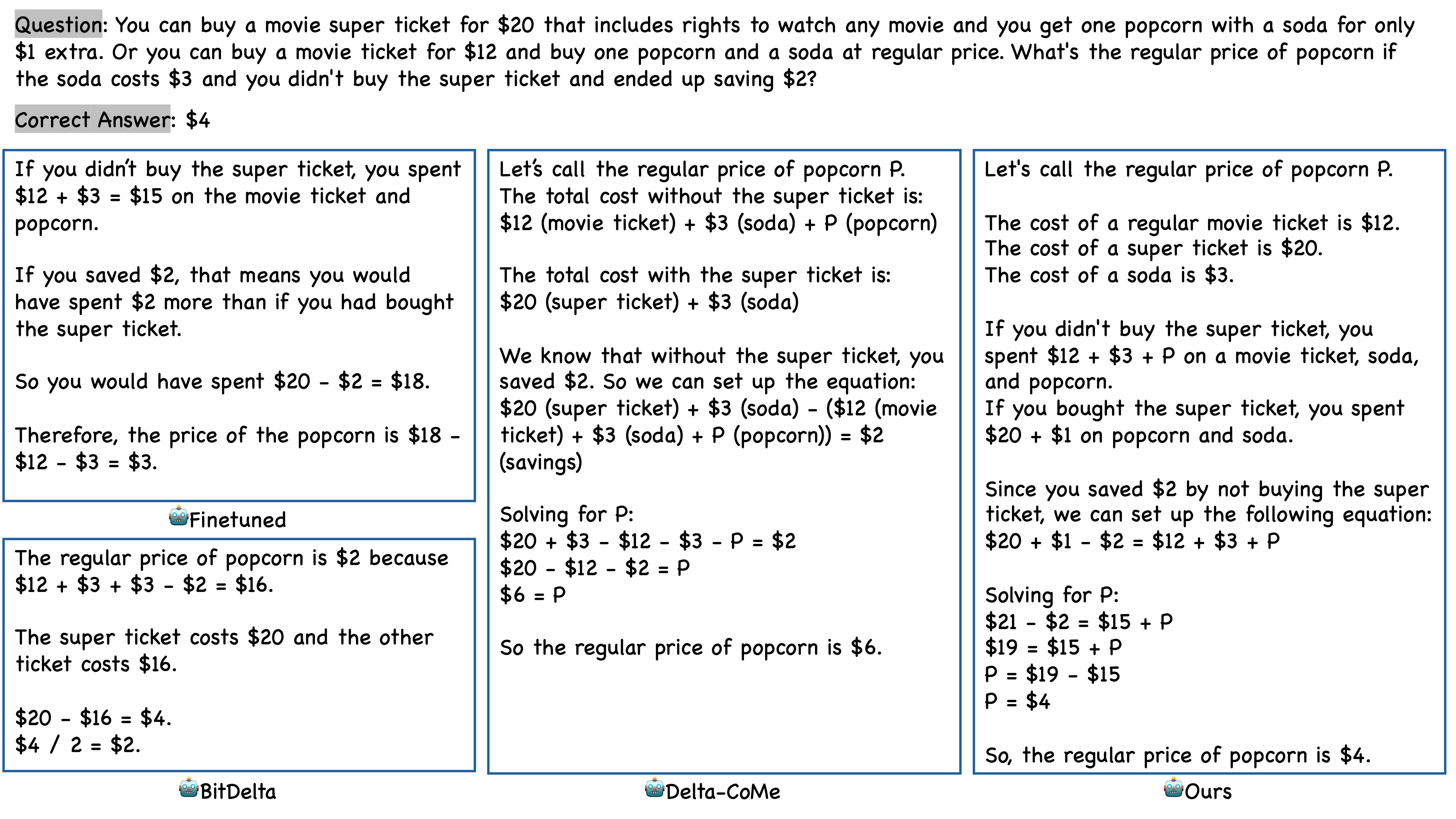}
    \caption{Case Study of a math task for different delta compression methods. Only our \methodshort{} outputs the right answer.
    }
    \label{fig:case_app}
\end{figure*}

\subsection{Experiments on eight RoBERTa-base models}
\label{sec:app-roberta}

\begin{table*}[t]

\centering
\caption{The performance of different methods on RoBERTa-base models.}
\label{tab:performance_roberta_base} 
\resizebox{0.95\textwidth}{!}{
\begin{tabular}{l|cccccccc|c}
\toprule

{Methods}& CoLA&SST2&MRPC&STSB&QQP&MNLI&QNLI&RTE & \textbf{Avg} \\
\midrule
{Finetuned}  & 60.18&94.04&89.22&90.63&91.41&87.20&92.71&79.06&86.48 \\
\midrule
{BitDelta}~\cite{liu2024bitdelta} & 36.83 & 93.12 & 88.73 &   84.82& 90.00&85.57&91.73 & 70.40 &80.15\\

Delta-CoMe~\cite{ping2024deltacome}&
60.20&	93.69 & 87.99& 90.53& 89.17 &82.66&92.57& 76.90 & 84.21
\\

\midrule

\textbf{\methodshort{} (Ours, p=16)} & 59.30&94.15 & 89.71 & 89.23 &90.84 & 86.48 & 92.62 & 76.90 &  84.90 
\\ 
\textbf{\methodshort{} (Ours, p=12)} & 58.80 & 94.27 & 89.22 & 89.74 & 91.08 & 87.00 & 92.44 & 77.26&84.98
\\

\bottomrule
\end{tabular}
}

\end{table*}

\noindent \textbf{Settings.} 
Following the setting from \cref{sec:exp_slm},
we validate the performance of RoBERTa-base~\cite{liu2019roberta} models on different tasks from GLUE~\cite{wang2018glue} benchmark, respectively CoLA~\cite{warstadt2019neural}, 
SST-2~\cite{socher2013recursive}, 
MRPC~\cite{dolan2005automatically},
STS-B~\cite{cer2017semeval},
QQP~\cite{iyer2017first},
MNLI~\cite{williams2017broad},
QNLI~\cite{rajpurkar2016squad}, and RTE~\cite{giampiccolo2007third}.
For RoBERTa-base models, CoLA is evaluated by the Matthews correlation
coefficient, STS-B is evaluated by the average value of Pearson and Spearman correlation coefficients, and the rest of the tasks are evaluated by exact match accuracy.

\noindent \textbf{Results.}
The results are shown in \cref{tab:performance_roberta_base}.
It can be seen that our \methodshort{} outperforms both baseline methods. 
The results further demonstrate the applicability of  \methodshort{} to relatively small language models.

\begin{table*}

\centering
\caption{Multi-task performance when merging ViT-B/16 models on eight tasks.}
\label{tab:performance_vitbase16} 
\resizebox{\textwidth}{!}{
\begin{tabular}{l|cccccccc|c}
\toprule

{Methods}& SUN397 & Cars &RESISC45 & EuroSAT & SVHN & GTSRB & MNIST & DTD  & \textbf{Avg Acc} \\
\midrule
Finetuned& 81.8&86.8&96.9&99.8&97.9 &99.2&99.8&82.1 &93.0
\\
\midrule
BitDelta~\cite{liu2024bitdelta}&81.2 &85.5 & 96.4 &99.5&97.6&98.8&99.6 & 81.3&92.5
\\
Delta-CoMe~\cite{ping2024deltacome}
&81.5 & 85.4 & 96.6 & 99.8 & 97.8 & 99.0 & 99.7 & 81.8 & 92.7

\\
\midrule
\textbf{\methodshort{} (Ours, p=16)}&81.8&86.4&96.3&99.8&97.5&99.0&99.7&82.1&92.8
\\
\textbf{\methodshort{} (Ours, p=8)}
&82.1&87.1 & 96.5 & 99.7&97.6 & 99.0 & 99.7&82.1&93.0
\\

\bottomrule
\end{tabular}
}

\end{table*}

\subsection{Experiments on eight ViT-B/16 models}
\label{sec:app-vit-b-16}
\noindent \textbf{Settings.} We follow the settings from \cref{sec:exp_vit} and apply ViT-B/16~\cite{dosovitskiy2020image}, which is a CLIP~\cite{clip} model's visual encoder, as the pre-trained model.
The performance of each method is evaluated by eight image classification tasks, including SUN397~\cite{xiao2010sun},
Cars~\cite{krause20133d},
RESISC45~\cite{cheng2017remote},
EuroSAT~\cite{helber2019eurosat},
SVHN~\cite{yuval2011reading},
GTSRB~\cite{stallkamp2011german},
MNIST~\cite{lecun1998mnist}, and
DTD~\cite{cimpoi2014describing}.
All the datasets are evaluated by accuracy.

\noindent \textbf{Results.} 
The experimental results are shown in \cref{tab:performance_vitbase16}.
Our \methodshort{} outperforms the baseline methods, BitDelta~\cite{liu2024bitdelta} and Delta-CoMe~\cite{ping2024deltacome}, and performs closely to the finetuned model. Under the patch-size setting of $p=8$, the average performance on the eight vision tasks equals the finetuned model.

\subsection{Additional Case Studies}
\label{sec:case_app}
We present an additional case study in \cref{fig:case_app}. 
The question is from GSM8K~\cite{cobbe2021trainingverifierssolvemath}, a math-solving dataset. We compare the answer from the finetuned model~\cite{numina_math_7b} and delta-compressed models by BitDelta~\cite{liu2024bitdelta}, Delta-CoMe~\cite{ping2024deltacome}, and our \methodshort{}. 
It can be seen that among all the methods, only our \methodshort{} outputs the correct answer, even outperforming the uncompressed finetuned model.

\section{Discussion: Why performance improves under partial experimental settings?}
\label{app:discuss_perf_increase}

Under partial experimental settings, the compressed model may even outperform the finetuned model.
The same phenomenon can be found in both merging and delta compression fields~\cite{yu2023language, huang2024emrmerging, ping2024deltacome}.
This may be due to the over-fitting or under-fitting of the finetuned models, that is, not the optimal 
position in the loss basin. 
Delta compression or merging with other models brings about the shift in the loss basin, thus resulting in a better performance.




 



\section{Limitations and future works}

Despite the convincing results, the proposed \methodshort{} suffers from several limitations. On the one hand, compared to existing methods, \methodshort{} requires a little additional storage overhead, which has been discussed in \cref{sec:overhead}.
On the other hand, \methodshort{} requires some computational resources and time to finish the DCT process.
Due to the DCT process of each patch and each layer is independent, the time cost caused by DCT can be significantly lowered by parallel computing or distributed computing.



Further improving the performance of delta compression and exploring the performance upper bound while lowering the compression costs is a significant direction for future work.


\end{document}